\documentclass[conference]{IEEEtran}
\IEEEoverridecommandlockouts
\usepackage{cite}
\usepackage{amsmath,amssymb,amsfonts}
\usepackage{algorithmic}
\usepackage{graphicx}
\usepackage{textcomp}
\usepackage{xcolor}
\usepackage{amsmath}
\usepackage{amsthm}
\usepackage{amssymb}
\usepackage{bm}
\usepackage{tikz}
\usepackage{multirow}
\usepackage{booktabs}
\usepackage{decimalcomma}
\usepackage{stfloats} 
\usetikzlibrary{positioning, arrows.meta}   

\graphicspath{ {./Figures/} }

\def\BibTeX{{\rm B\kern-.05em{\sc i\kern-.025em b}\kern-.08em
    T\kern-.1667em\lower.7ex\hbox{E}\kern-.125emX}}

\newcommand{\rf}{\mathbb R}

\begin{document}

\title{Subdomain-aware representation compression for pretrained image embeddings}

\author{
\IEEEauthorblockN{Poowanut Niamluang}
\IEEEauthorblockA{
\textit{Department of Computer Engineering} \\
\textit{Kasetsart University}, Bangkok, Thailand \\
\texttt{poowanut.nice@gmail.com}}
\and
\IEEEauthorblockN{Jittat Fakcharoenphol}
\IEEEauthorblockA{\textit{Department of Computer Engineering} \\
\textit{Kasetsart University}, Bangkok, Thailand \\
\texttt{jittat@gmail.com}}
}

\maketitle

\begin{abstract}
Dimensionality reduction is a well-known technique for improving space efficiency, typically applied uniformly across an entire dataset.
This paper investigates the possibilities of using dimensionality reduction techniques for subdomain representation compression.  
We explore standard techniques such as Principal Component Analysis (PCA) and Linear discriminant analysis (LDA) in image domains.  
The results not only demonstrate the expected improvements in space and computation complexity crucial for edge-device ML applications but also show improvements in accuracy over direct full-embedding procedure.
One possible explanation is that dimensionality reduction effectively extracts subdomain features.
We also performed experiments to demonstrate transfer learning capabilities using the compressed representations.
\end{abstract}


\section{Introduction}

Pretrained embeddings are one of the key building blocks of modern ML applications as they enable the transformation from the raw data space to a more manageable and informative feature space.
Typically, the dimensions of the embedding spaces are designed to capture the entire distribution of the domain data, resulting in a high-dimensional representation that may include irrelevant or redundant information for the specific task at hand.

In this work, we present a preliminary investigation on the potential usage of standard dimensionality reduction techniques such as PCA~\cite{AbdiW10-pca} and LDA~\cite{Fisher1936LDA, Hastie2009elements} for subdomain representation compression of pretrained embeddings.  Obviously, reducing the dimension reduces the resource requirements, but it is a conventional wisdom that removing information hurts the quality of the representation.
However, when the focus is the subdomain, it is possible that dimensionality reduction procedures can pick out important features and enhance the representation's quality for the specific task.  
We note that there are many theoretical results supporting this idea, e.g., results on spectral clustering~\cite{VEMPALA2004841} and on learning linear subspaces~\cite{pmlr-v139-tripuraneni21a}.

We perform experiments on the image domains using clustering as a proxy task to show the effectiveness of standard dimensionality reduction procedures on subdomain representation compression under two settings: (i) an unsupervised setting where only samples of subdomain are used to train the compression scheme and (ii) a supervised setting where labeled data are also available.  
To also study whether compression can choose important features, we include a zero-shot transfer learning setting where the compression procedure is trained on a different but related subdomain.

Our procedure uses sampled embeddings (maybe with labels) from the subdomain to learn the compression transformation, 
and given an embedding of an actual data point, it can produce a compressed representation.  Note that this procedure requires access to the original embedding model to obtain the full embedding of a data point before it can perform the compression.

\subsection{Applications}

While dimensionality reduction has wide applications and this paper focuses mostly on properties of the compression for subdomain representation, there are many potential applications under our exact formulation of the stated compression procedure.

Consider a setting where access to the full embedding model is constrained (e.g., remote area), and the user operates on a small edge device. The user can collect data without class annotations, obtain their full embeddings remotely, and apply the proposed compression procedure, storing only the compressed embeddings locally. 
After this step, remote access to the full embedding model can be eliminated.
The collected data and their corresponding compressed embeddings can then be used to distill a lightweight local encoder that approximates the mapping from raw inputs to compressed embeddings. This enables the  generation of compressed embeddings for newly observed samples using only local computation. Downstream inference can be performed without direct access to the original embedding model.
For example, the user might later learn about the classes of the collected data, and with the collected data, the user can perform clustering analysis.

There are other situations where a user with low computational power may want to perform ML tasks without relying on large remote ML services.  When the class labels are of high value, the user may not want to reveal the labels to remote servers.  Obtaining the full embeddings remotely and performing ML tasks on compressed embeddings locally might provide privacy preserving ML procedure in this case.

\subsection{Our contributions}

We focus on using standard dimensionality reduction procedures to compress the representation using sampled subdomain raw data.
While our approach is straight-forward, the results from our experiments based on 3 datasets with 2 pretrained embedding foundational models show that this process provides two important benefits simultaneously: improved space efficiency and enhanced downstream task performance.  We also observed that compression performance can be significantly better than typical usage (of $50\%$~\cite{RaunakGM2019-effective-dim-reduction-word-embeddings}) and with supervised compression, the compression ratio can be significantly lower.  We also consider transfer learning capabilities of the compressed representations.

\subsection{Related work}

Dimensionality reduction is a widely used technique.  Raunak, Gupta, and Metze~\cite{RaunakGM2019-effective-dim-reduction-word-embeddings} studied the applications to pretrained word embeddings, while May~{\em et al.}~\cite{MayZDR2019-downstream-compressed-word-embeddings} investigated compressed word embeddings for downstream tasks.  Ma~{\em et al.}~\cite{MaLSXL2021-emnlp-compress-retrieval} also used PCA to improve space efficiency in retrieval.

Dealing with subdomain representations is also referred to as subdomain adaptation.  While preparing this manuscript, we encountered an independent work of Zuo and Khashabi~\cite{ZuoK2026-more-than-efficiency} (the arXiv version appeared on 20 January 2026) that also uses PCA and investigates embedding compression for domain adaptation in dense retrieval for textual data.  
The main difference to our work is that their paper focuses on textual retrieval tasks.
Also, while their findings support our work, the lower dimensional representations only moderately dominate the full dimensional representation.  In our case on image domains, the improvements from compressed embeddings are more observable.

Most domain adaptation research (e.g.,~\cite{margaritis2025efficientdomainadaptationmultimodal,barnes-etal-2018-projecting,JMLR:v17:15-207}) focused on improving quality. Notable work on domain adaptation that focuses on space efficiency is by Venkateswara~{\em et al.}\cite{Venkateswara2017-cvpr-deep-hashing-network}.
Transfer learning is an important topic in ML, but can also be considered as a version of domain adaptation problems.  For image domain Kornblith~{\em et al.}~\cite{KornblithSS2019-do-better-imagenet-transfer} investigated the possibility of using good representations learned from one domain to improve performance in another domain.

There are theoretical results explaining and investigating the problem of learning subspace representation (see, e.g.,~\cite{pmlr-v139-tripuraneni21a,ArpitNIPS2014-dimreduction-subspace}).

\section{Problem statement}

We first state a general formulation based on dimensionality reduction problems.
Given a set of $n$ data points with their embeddings $e_1,\ldots, e_n$ in $d$-dimensional space $\rf^d$, our goal is to find a transformation $f: \rf^d \to \rf^k$ where $k < d$, such that $e'_i=f(e_i)$ preserves the important features from the original embedding.  
Moreover, in our case, the original embeddings $e_i$'s are mapping from the data domain $\mathcal D$ while the actual data points are sampled from the subdomain $\mathcal D'\subseteq \mathcal D$.  
To compute $f$, we assume that we have $m$ sampled data points with their embeddings for the unsupervised compression.  In some rare case, we may also have their class labels for the supervised compression.

While we believe that our approach works well under various domains and for many downstream tasks, we focus on an image domain and clustering problems.  
Therefore, in this preliminary work, during the training phase, we are given a set of $m$ images from subdomain $\mathcal D'$ with their embeddings $e_1,\ldots, e_m$ in $d$-dimensional space $\rf^d$ based on a pretrained image embedding model over domain $\mathcal D$, with possibly their labels.  
We would like to compute a transformation $f: \rf^d \to \rf^k$ where $k < d$, such that $e'_i=f(e_i)$ preserves the important features from the original embedding for clustering tasks for the actual $n$ data points. 

We are also interested in transfer learning of the compressed representation.  
Specifically in this case, the sampled $m$ images are obtained from domain $\mathcal D''$ while the actual data points are from domain $\mathcal D'\neq\mathcal D''$.  Both domains might share some important characteristics.
We experimented on a variety of source domains to see if the source-target similarity matters. 

\section{Preliminary}

\subsection{Embeddings and pretrained embeddings}

Data representation using feature preserving embeddings~\cite{Mikolov-NIPS2013-word-embeddings, pennington-etal-2014-glove, devlin2019bertpretrainingdeepbidirectional,oquab2024dinov2learningrobustvisual} is a key part of ML development.  
Embedding techniques transform raw data (images, texts, or audio) into vector representations.  
This process enables efficient storage, retrieval, and processing.  In most cases, good embeddings have to be learned from a large pool of data and take a huge amount of effort to compute and optimize.  
Pretrained embeddings based on large-scale collected corpuses such as word2vec~\cite{Mikolov-NIPS2013-word-embeddings}, Glove~\cite{pennington-etal-2014-glove}, BERT~\cite{devlin2019bertpretrainingdeepbidirectional}, Vision Transformer~\cite{dosovitskiy2021imageworth16x16words}, DINOv2~\cite{oquab2024dinov2learningrobustvisual}, and CLIP~\cite{radford2021learningtransferablevisualmodels} enable efficient transfer learning and adaptation to new tasks with minimal additional training.

In this work, we focus on the image domain and use pretrained image embeddings such as DINOv2~\cite{oquab2024dinov2learningrobustvisual} and CLIP~\cite{radford2021learningtransferablevisualmodels}.  DINOv2~\cite{oquab2024dinov2learningrobustvisual} uses Vision Transformers to learn robust image embeddings via self-supervised learning, excelling in tasks such as clustering and classification.  CLIP~\cite{radford2021learningtransferablevisualmodels} aligns image-text pairs in a shared space, enabling the production of embeddings that support zero-shot transfer and cross-modal understanding for diverse tasks.

\subsection{Dimensionality reduction}

Dimensionality reduction techniques aim to project high-dimensional data into a lower-dimensional space while preserving essential information. These methods are crucial for mitigating the "curse of dimensionality," reducing computational complexity, and improving model interpretability. Dimensionality reduction can be broadly categorized into unsupervised methods, which rely on intrinsic data properties, and supervised methods, which incorporate label information to enhance task-specific performance.

More specifically, we are given a dataset from a $n$-dimensional space $\mathbb{R}^d$. Our goal is to project the data onto a $k$-dimensional subspace $\mathbb{R}^k$ ($k < d$) using dimensionality reduction techniques.

\textbf{Principal Component Analysis (PCA)~\cite{AbdiW10-pca}} is an unsupervised dimensionality reduction technique that projects data from a high-dimensional space to a lower-dimensional subspace while preserving maximum variance. Given a dataset $\mathbf{X} \in \mathbb{R}^{m \times d}$, PCA computes the covariance matrix, performs eigenvalue decomposition, and selects the top $m$ eigenvectors corresponding to the largest eigenvalues to form the projection matrix $\mathbf{W} \in \mathbb{R}^{m \times k}$. The reduced data is obtained as $\mathbf{X}_{\text{reduced}} = \mathbf{X}_{\text{centered}} \mathbf{W}$. PCA ensures that the projected data retains the most significant variance, making it effective for preprocessing and noise reduction.

\textbf{Linear Discriminant Analysis (LDA)~\cite{Fisher1936LDA, Hastie2009elements}} is a supervised dimensionality reduction technique that projects data to a lower-dimensional subspace while maximizing class separability. Given a dataset $\mathbf{X} \in \mathbb{R}^{m \times d}$ with corresponding class labels $\mathbf{Y} \in \mathbb{R}^m$, LDA computes the within-class scatter matrix $\mathbf{S}_W$ and the between-class scatter matrix $\mathbf{S}_B$. The projection matrix $\mathbf{W} \in \mathbb{R}^{m \times k}$ is obtained by solving the generalized eigenvalue problem $\mathbf{S}_B \mathbf{w} = \lambda \mathbf{S}_W \mathbf{w}$, where $\lambda$ represents the eigenvalues. The top $m$ eigenvectors corresponding to the largest eigenvalues form $\mathbf{W}$. The reduced data is then obtained as $\mathbf{X}_{\text{reduced}} = \mathbf{X} \mathbf{W}$. LDA ensures that the projected data maximizes class separability, making it highly effective for classification tasks.

\section{Our methodology}

We apply standard dimensionality reduction techniques such as PCA and LDA to compress the representations of subdomains.

{\em Unsupervised compression.} We are given a sampled data set of $m$ images.  We use pretrained embeddings to obtain the embedding $\mathbf{X} \in \mathbb{R}^{m \times d}$, where $d$ is the dimensionality of the embedding space.  
We then apply PCA to produce the projection matrix $\mathbf{W}\in \mathbb{R}^{d \times k}$, where $k < d$. 
For each actual data point $x_i$ with its embedding $e_i \in \mathbb{R}^d$, we compute the compressed representation as $e'_i = e_i \mathbf{W}\in\mathbb{R}^k$.

{\em Supervised compression.} When the labels of the sampled data are available, we can use LDA to produce a projection matrix $\mathbf{W}\in \mathbb{R}^{d \times k}$, where $k < d$. 

We use clustering problems as test cases to evaluate the quality of the compressed representations.  In our implementation, we employ both the k-Means algorithm~\cite{Lloyd1982least} and its minibatch variant~\cite{Sculley2010web-minibatch-kmeans}.

\section{Experiments}

\subsection{Datasets, metrics, and models}
We use standard benchmarks derived from the ILSVRC-2012 (ImageNet-1k)~\cite{imagenet15russakovsky} dataset to evaluate clustering performance across different downstream tasks as follows.

\begin{itemize}
    \item \textbf{ImageNet-10}: A coarse-grained subset of ImageNet-1k containing $10$ distinct classes, each with $1,300$ images. This dataset serves as a baseline for measuring clustering performance on visually distinct semantic categories.
    \item \textbf{ImageNet-Dog-15}: A fine-grained subset consisting of $15$ different dog breeds (each with $1,300$ images) selected from ImageNet-1k. This task evaluates the model's ability to distinguish between semantically similar sub-categories, posing a greater challenge for subspace separability.
    \item \textbf{TinyImageNet}: A large-scale dataset with $200$ classes and images resized to $64 \times 64$ pixels. It provides a more diverse and computationally demanding benchmark to assess the robustness of post-processing methods across a wider semantic range. 
    \item \textbf{Domain Transfer Splits (Source domains vs. Cats)}: For transfer learning experiments, we curate specific subsets from ImageNet-1k based on WordNet synset IDs. The \textit{Source Domains} consist of Dogs ($\sim 2,000$ samples, $118$ classes), 
    Bird ($\sim 2,000$ samples, $142$ classes), 
    Household ($\sim 2,000$ samples, $100$ classes), and 
    Vehicle ($\sim 2,000$ samples, $101$ classes).  
    The \textit{Target Domain} consists of Cat (labels 281--293), comprising approximately $500$ samples, with $13$ classes.
\end{itemize}

To evaluate the quality of the clustering, we use the following standard metrics:
\begin{itemize}
    \item \textbf{V-Measure (V)}: Combines homogeneity and completeness into a single metric, where homogeneity ensures that clusters contain only samples from a single class, and completeness ensures that all samples of a class are grouped into the same cluster.
    \item \textbf{Normalized Mutual Information (NMI)}: Quantifies the dependency between predicted cluster labels and true labels, normalized to produce values within the range [0, 1].
    \item \textbf{Adjusted Rand Index (ARI)}: Measures the agreement between predicted and true cluster assignments by evaluating all sample pairs, adjusting for random chance.
    \item \textbf{Accuracy}: Measures the proportion of correctly clustered samples after optimal label mapping using the Hungarian algorithm.
\end{itemize}

Experiments are conducted using two foundation model backbones:
\begin{enumerate}
    \item \textbf{DINOv2 (ViT-B/14)}~\cite{oquab2024dinov2learningrobustvisual}: Embedding dimension $d=768$.
    \item \textbf{CLIP (ViT-B/32)}~\cite{radford2021learningtransferablevisualmodels}: Embedding dimension $d=512$.
\end{enumerate}

For each configuration, we run the experiment 5 times and present the average results.

\subsection{Unsupervised Compression}
\begin{figure*}[t]
    \centering
    \includegraphics[width=\textwidth]{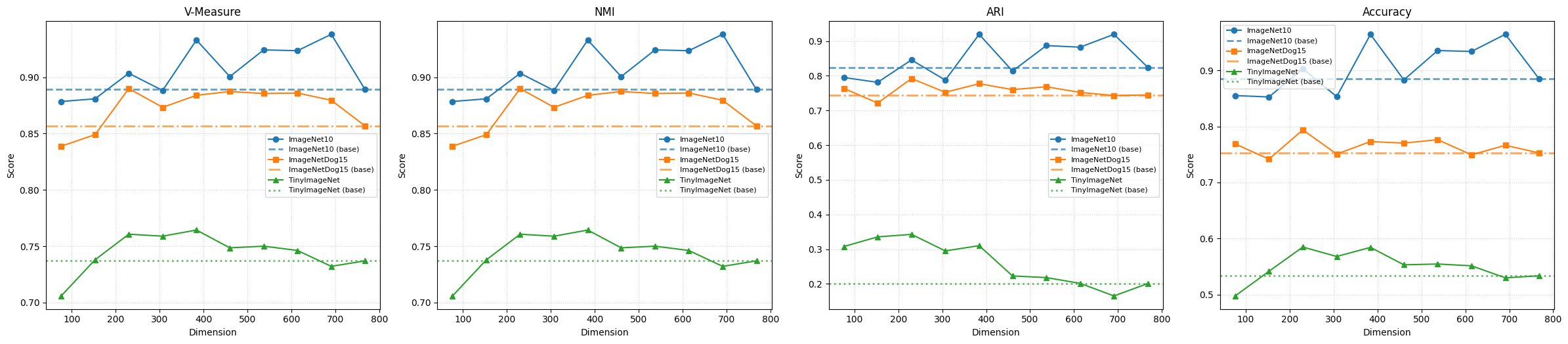}
    \caption{PCA clustering score on ImageNet-10, ImageNetDog-15, and TinyImageNet using DINOV2-base ($d=768$).}
    \label{fig:pca-dino}
\end{figure*}

\begin{figure*}[t]
    \centering
    \includegraphics[width=\textwidth]{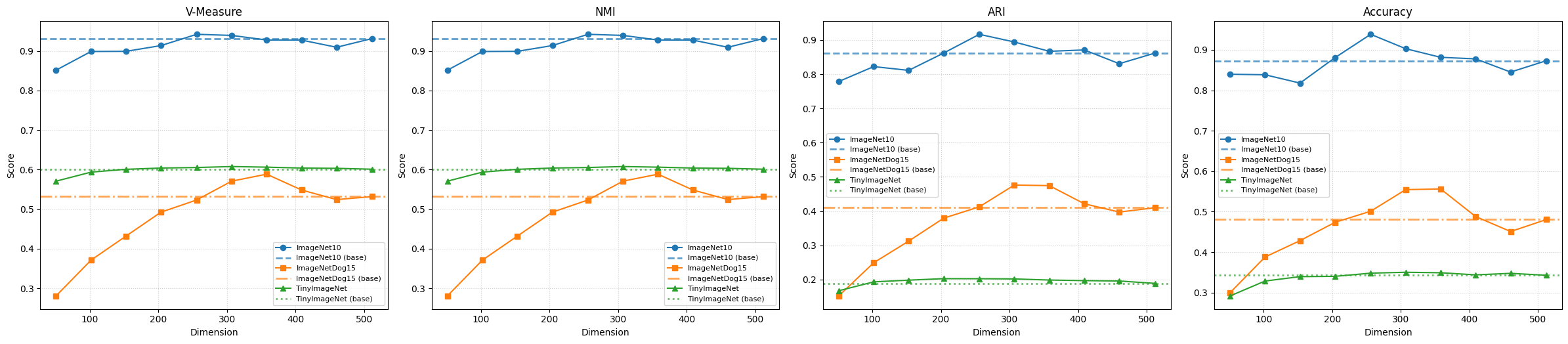}
    \caption{PCA clustering score on ImageNet-10, ImageNetDog-15, and TinyImageNet using CLIP ViT-B/32 ($d=512$).}
    \label{fig:pca-clip}
\end{figure*}

For each benchmark (ImageNet-10, ImageNet-Dog-15, and TinyImageNet), we take $5000$ sampled images with their embedding from one of the foundation models to train the PCA projection matrix $\mathbf{W}$ that reduced the dimension down to various parameters $k$.  
The projection is used to generate compressed embeddings for the rest of the input images, which are clustered using the k-Means algorithm.  The results are shown with the base line performance of the full-embedding case in Figures~\ref{fig:pca-dino} and~\ref{fig:pca-clip}.

In most cases, we observe improvements in clustering quality when using the projected embeddings.  
We would like to emphasize that typically, as mentioned in~\cite{RaunakGM2019-effective-dim-reduction-word-embeddings}, the reduced dimension should be at least 50\% of the original dimension, however in the case of subdomain compression, in many datasets notably in ImageNetDog-15 and TinyImageNet (with DINOv2) and in TinyImageNet (with CLIP), we can see improvements even when the reduced dimension is 25\% or lower of the original. 

\subsection{Supervised Compression}
\begin{figure*}[ht]
    \centering
    \includegraphics[width=\textwidth]{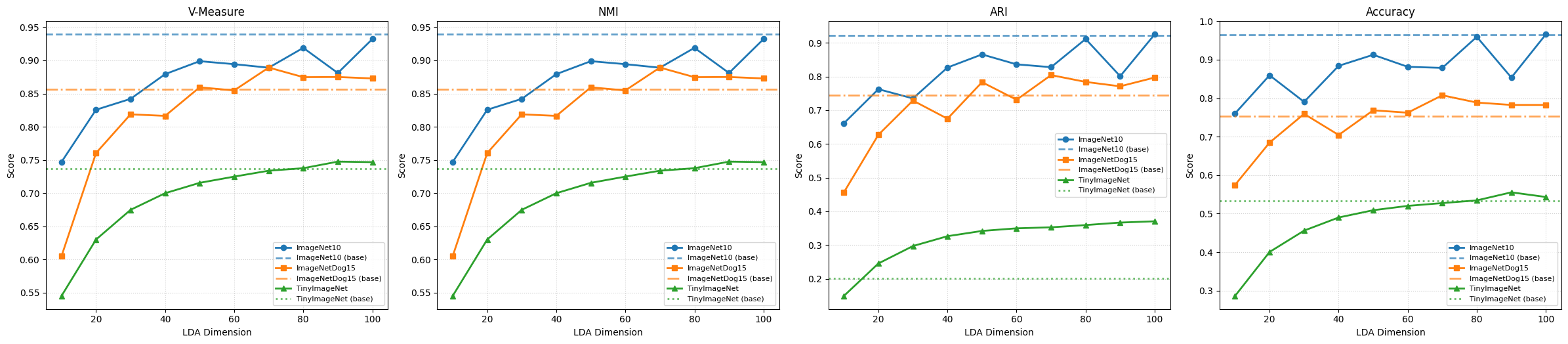}
    \caption{LDA clustering score on ImageNet-10, ImageNetDog-15, and TinyImageNet  using DINOv2-base ($d=768$).}
    \label{fig:dino-lda}
\end{figure*}

\begin{figure*}[ht]
    \centering
    \includegraphics[width=\textwidth]{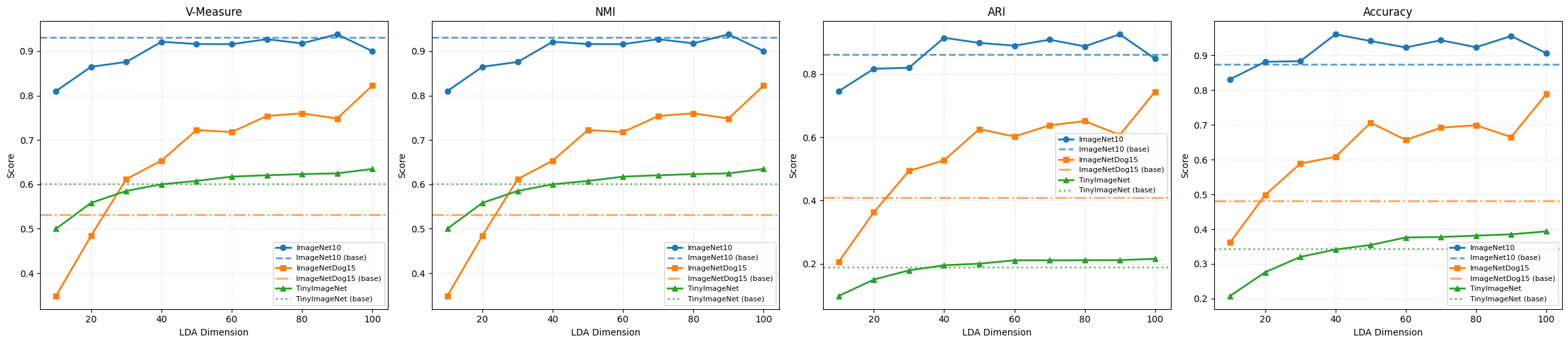}
    \caption{LDA clustering score on ImageNet-10, ImageNetDog-15, and TinyImageNet  using CLIP ViT-B/32 ($d=512$).}
    \label{fig:clip-lda}
\end{figure*}

For supervised compression, we also incorporate image labels and use LDA to compute the projection matrix $\mathbf{W}$.  
Since LDA only allows the number of dimensions to be at most the number of classes, the value of $k$ is of limited range (compared to previous results with PCA).
The results are shown in Figures~\ref{fig:dino-lda} and~\ref{fig:clip-lda}.


When the labels are available, the compression ratio with competitive performance against the full embeddings can be much lower than in the PCA cases.  For example, in ImageNetDog-15 and TinyImageNet (for DINOv2) and TinyImageNet (for CLIP), the compression ratio can be as low as 10\%, and in ImageNetDog-15 (for CLIP) the ratio can be as low as 5\%.

\begin{table}[!tb]
\centering
\caption{Best clustering performance using CLIP.\\ Dimensionality $k$ is shown in parentheses.}
\label{tab:clip_results}
\footnotesize
\setlength{\tabcolsep}{3pt}
\begin{tabular}{l l c c c}
\toprule
Dataset & Method ($k$) & NMI & ARI & ACC \\
\midrule
\multirow{4}{*}{ImageNet-10}
 & Full ($512$) & 0.9314 & 0.8618 & 0.8732 \\
 & LDA ($90$) & 0.9381 & \textbf{0.9260} & \textbf{0.9549} \\
 & PCA ($256$) & \textbf{0.9420} & 0.9167 & 0.9385 \\
\midrule
\multirow{4}{*}{ImageNet-Dog}
 & Full ($512$) & 0.5318 & 0.4099 & 0.4805 \\
 & LDA ($100$) & \textbf{0.8224} & \textbf{0.7441} & \textbf{0.7881} \\
 & PCA ($358$) & 0.5885 & 0.4744 & 0.5561 \\
\midrule
\multirow{4}{*}{TinyImageNet}
 & Full ($512$) & 0.6010 & 0.1888 & 0.3428 \\
 & LDA ($100$) & \textbf{0.6347} & \textbf{0.2155} & \textbf{0.3935} \\
 & PCA ($307$) & 0.6077 & 0.2017 & 0.3502 \\
\bottomrule
\end{tabular}
\end{table}

\begin{table}[!tb]
\centering
\caption{Best clustering performance using DINOv2. \\ Dimensionality $k$ is shown in parentheses.}
\label{tab:dino_results}
\footnotesize
\setlength{\tabcolsep}{3pt}
\begin{tabular}{l l c c c}
\toprule
Dataset & Method ($k$) & NMI & ARI & ACC \\
\midrule
\multirow{3}{*}{ImageNet-10}
 & Full ($768$) & 0.8890 & 0.8244 & 0.8850 \\
 & LDA ($100$) & 0.9321 & \textbf{0.9252} & \textbf{0.9661} \\
 & PCA ($691$) & \textbf{0.9381} & 0.9196 & 0.9646 \\
\midrule
\multirow{3}{*}{ImageNet-Dog}
 & Full ($768$) & 0.8567 & 0.7450 & 0.7528 \\
 & LDA ($70$) & 0.8892 & \textbf{0.8042} & \textbf{0.8076} \\
 & PCA ($230$) & \textbf{0.8899} & 0.7921 & 0.7941 \\
\midrule
\multirow{3}{*}{TinyImageNet}
 & Full ($768$) & 0.7370 & 0.2012 & 0.5334 \\
 & LDA ($90$) & 0.7475 & \textbf{0.3666} & 0.5554 \\
 & PCA ($384$) & \textbf{0.7644} & 0.3103 & \textbf{0.5841} \\
\bottomrule
\end{tabular}
\end{table}

Tables~\ref{tab:clip_results} and~\ref{tab:dino_results} summarize the best clustering performance achieved for each dataset and method using CLIP and DINOv2, respectively, with the corresponding embedding dimensionality $k$ reported in parentheses. 
Overall, the results indicate that dimensionality reduction techniques, including PCA LDA are able to retain and in several cases even improve clustering performance when compared with the original embedding.

\subsection{Transfer learning}
To investigate the generalization of discriminative subspaces across semantically related domains, we implemented a \textbf{Zero-Shot Subspace Transfer}. 

We hypothesize that the geometric directions that optimally separate fine-grained classes in a source domain (e.g., dog) capture fundamental "features of difference" (such as shape, texture, or pose) that remain relevant for separating classes in a distinct but related target domains (e.g., cat), even without training on the target domain.  To test this hypothesis, we performed experiments on transfer learning to the target domain from a variety of source domains with different degrees of similarity.

We work on many disjoint domains from ImageNet-1k:
\begin{itemize}
    \item \textbf{Four Source Domains ($D_{S_i}$)}: 
    ``Dog'' ($D_{S_1}$, Synsets 151--268) with $N_{S_1}=2000$ samples, 
    ``Bird'' ($D_{S_2}$, Synsets 7--148) with $N_{S_2}=2000$ samples,
    ``Household'' ($D_{S_3}$, Synsets 801--900) with $N_{S_3}=2000$ samples, and
    ``Vehicle'' ($D_{S_4}$, Synsets 650--750) with $N_{S_4}=2000$ samples.
    \item \textbf{Target Domain ($D_T$)}:  ``Cat'' (Synsets 281--293). \\ $N_T=500$ samples.
\end{itemize}

For each source domain $D_{S_i}$, the procedure is as follows:
\begin{enumerate}
    \item Feature Extraction: We extract fixed embeddings $\mathbf{X}_S$ and $\mathbf{X}_T$ using the frozen CLIP ViT-B/32 backbone (with $d=512$).
    \item Subspace Learning: We fit a projection matrix $W$ solely on the source domain $\mathbf{X}_S$. We evaluate two strategies:
    \begin{itemize}
        \item Supervised Transfer (LDA): $W$ is learned to maximize the separation of classes on $D_{S_i}$.
        \item Unsupervised Transfer (PCA): $W$ is learned to maximize the variance of features of $D_{S_i}$.
    \end{itemize}
    \item Zero-Shot Application: The learned projection $W$ is applied to the unseen target embeddings: $\mathbf{X}'_T = \mathbf{X}_T W$.
    \item Evaluation: We perform k-Means clustering on the transformed target embeddings $\mathbf{X}'_T$ and compare performance against the raw embeddings $\mathbf{X}_T$.
\end{enumerate}

\begin{figure*}[t]
    \centering
    \includegraphics[width=.8\columnwidth]{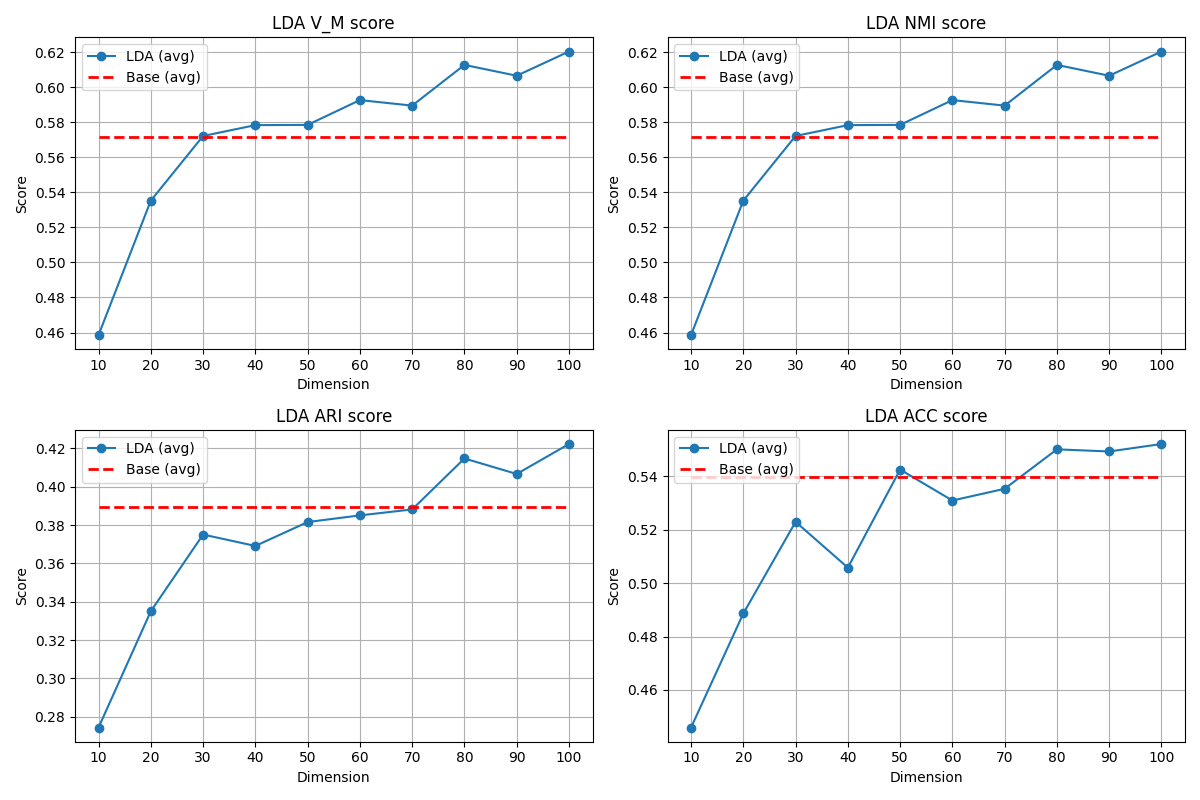}
    \ \ 
    \includegraphics[width=.8\columnwidth]{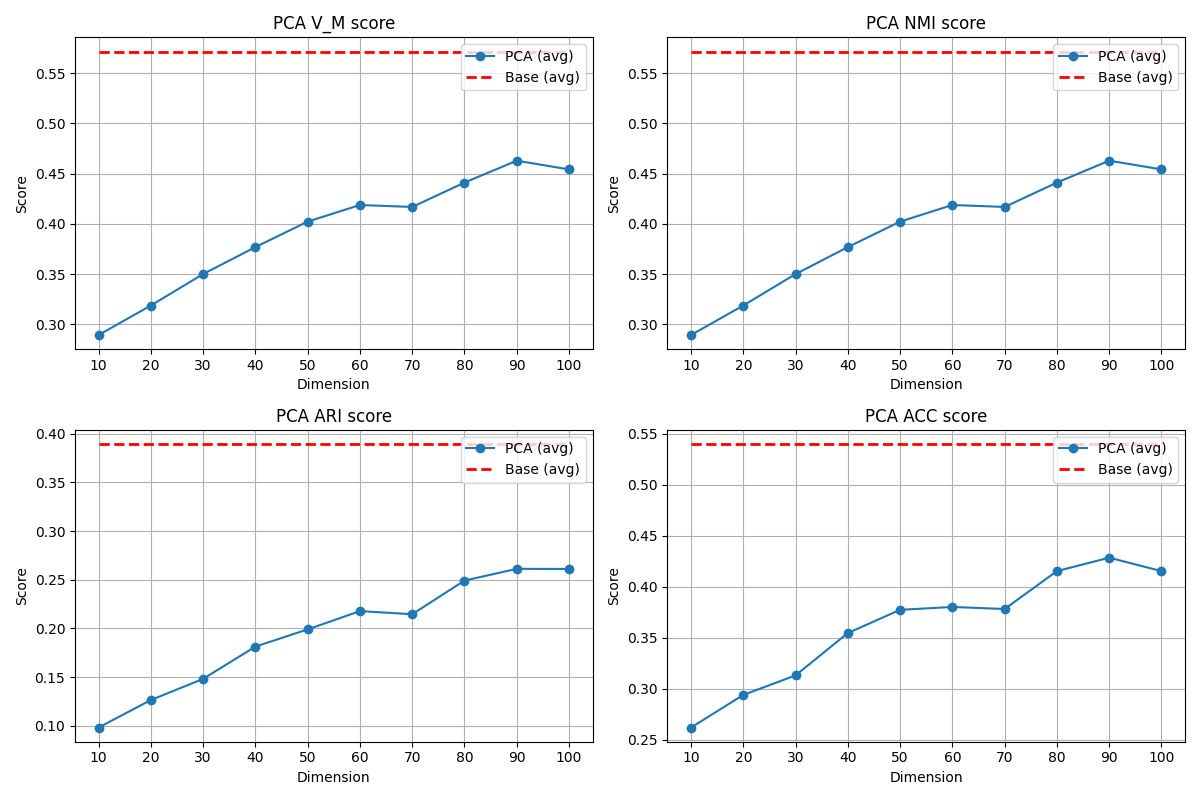}
    \caption{LDA and PCA transfer compression result for CLIP ($d=512$)}
    \label{fig:dog2cat}
\end{figure*}

Figure~\ref{fig:dog2cat} shows the clustering performance in the target domain using learned projection from Dog domain ($D_{S_1}$).  Note that while PCA-based compression shows slightly inferior results, but we only consider the dimension up to $100$ which is only $20\%$ of the original dimension $d=512$.
While PCA-based compression does not show any improvement, the LDA-based supervised compression even gives a significant boost in performance compared to the based line full-embedding.

\begin{figure*}[t]
    \centering
    \includegraphics[width=1.5\columnwidth]{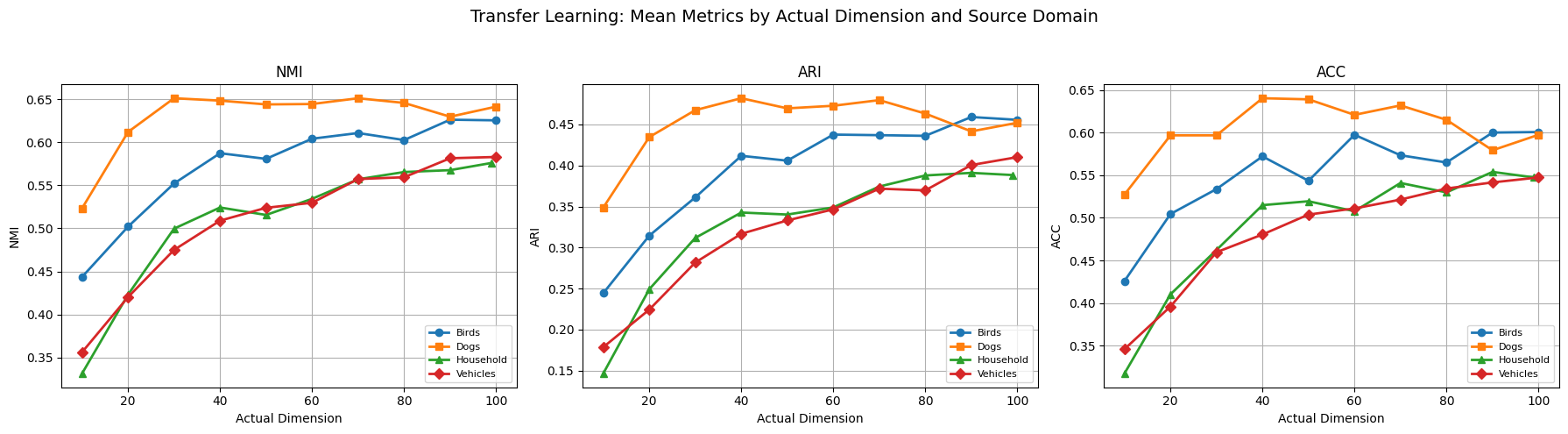}
    \ \ 
    \caption{LDA transfer compression result comparison for CLIP ($d=512$)}
    \label{fig:n2cat}
\end{figure*}

Figure~\ref{fig:n2cat} compares the performances between various source domains.  We note that the learned projection from Dog domain shows better overall results, following by the Bird domain.  The Household and Vehicle domains perform worse.  This experiment strongly hints that the learned projections focus on ``subdomain'' features, and this is the reason why transfer learning for related domains yields better results.

\section{Conclusion and future work}

This preliminary work demonstrates potential applications of dimensionality reduction to subdomain representation compression.
It is important to consider other downstream tasks, notably classification tasks.
Moreover, it remained to investigate the reason why compression improves the performance.  
While we hypothesize that the key condition is that the new domain is a subdomain of the original input space of the pretrained embeddings, we have not demonstrated decisively that this is the case.  This would be the most crucial follow-up work.

\bibliographystyle{IEEEtran}
\bibliography{compression}

\end{document}